\documentclass{Interspeech2024}
\usepackage{caption}
\usepackage{subcaption}
\usepackage{graphicx}
\usepackage[textwidth=0.6in]{todonotes}
\usepackage{hyperref}
\usepackage{booktabs}
\usepackage{amssymb}
\usepackage{amsmath}
\usepackage{multirow}

\interspeechcameraready

\title{ On the Encoding of Gender in Transformer-based ASR Representations }

\name{Aravind Krishnan, Badr M. Abdullah, Dietrich Klakow}

\address{Spoken Language Systems Group, Saarland University, Germany}
\email{\{akrishnan|babdullah|dietrich\}@lsv.uni-saarland.de}

\keywords{speech recognition, gender, linear erasure}

\begin{document}

\maketitle

\begin{abstract}
    
While existing literature relies on performance differences to uncover gender biases in ASR models, a deeper analysis is essential to understand how gender is encoded and utilized during transcript generation. This work investigates the encoding and utilization of gender in the latent representations of two transformer-based ASR models, Wav2Vec2 and HuBERT. Using linear erasure, we demonstrate the feasibility of removing gender information from each layer of an ASR model  and show that such an intervention has minimal impacts on the ASR performance. Additionally, our analysis reveals a concentration of gender information within the first and last frames in the final layers, explaining the ease of erasing gender in these layers. Our findings suggest the prospect of creating gender-neutral embeddings that can be integrated into ASR frameworks without compromising their efficacy.
\end{abstract}

\section{Introduction}

The advent of transformer-backed speech models like Wav2Vec2~\cite{baevski2020wav2vec}, HuBERT~\cite{hsu2021HuBERT}, WavLM~\cite{chen2022wavlm} and Whisper~\cite{radford2023whisper} has had a dramatic effect on the automatic speech recognition (ASR) paradigm, with impressive results on several benchmarks. Although we marvel at their "performance", recent research has questioned the inclusivity of ASR models, holding them accountable for propagating (mostly) human biases. One direction in uncovering model bias is on the basis of socio-linguistic attributes (See \cite{feng2024towards} for an overview). Gender has been a very popular attribute among them, with several works suggesting a bias towards male~\cite{garnerin2019gender} \textit{or} female~\cite{attanasio2024multilingual,fuckner2023uncovering,adda2005speech,tatman2017gender,alsharhan2020investigating} speakers. Apart from ASR models, these trends have also been reported in speech  translators and speech classifiers~\cite{savoldi2022under,costa2020evaluating, gorrostieta2019gender}.

Most works that study how gender influences a speech model's prediction use performance disparities to scrutinize the system's response to the speaker's gender. While this acts as a litmus test for detecting bias, more analysis is needed to understand \textit{how} these models encode and utilize gender in the first place. The closest work that explores this direction is \cite{attanasio2024multilingual}, where the authors probe for and discover gender embedded in the final layer of Whisper's encoder. In this work, we study the encoding and utilization of gender across the layers of two transformer-based  ASR models.  We expand our investigation by exploring if the detection of an attribute in the hidden layers of a model ensures its utilization by the model. In amnesic probing \cite{elazar2021amnesic}, the authors assert that utilization can only be guaranteed by elimination, i.e., Removing a property the model uses will disrupt it more than removing one it doesn't use. The authors show that the removal of semantic and syntactic properties affect BERT~\cite{devlin2018bert} differently. For removing an attribute from an embedding space, they employ Iterative Null-space Projection (INLP) \cite{ravfogel2020null}, where data points are projected onto the null space of the linear boundary that separates them into the classes of the target property.\\
This work proposes an experimental approach based on amnesic probing to study the encoding of gender in ASR models. Formally, we ask the following question: ``\textit{\textbf{How much gender-related information can we linearly erase from an ASR model and what are the implications of linear erasure?}}''. Following~\cite{ravfogel2020null,elazar2021amnesic, belrose2024leace}, we confine ourselves to the linear domain, probing and removing linearly encoded gender from the model's hidden layer representations.  To the best of our knowledge, this is the first work that employs linear erasure to analyze and anonymize ASR embeddings. 
Our contributions are as follows:
\begin{enumerate}
    \item We effectively remove linearly encoded gender information across all layers of the Wav2Vec2 and HuBERT models, without necessitating model training (\S{\ref{sec:tracking_erasure}}).
    \item We analyze the effects of linear erasure on the downstream ASR performance of the models (\S{\ref{sec:downstream_asr}}).
    \item We uncover a frame-level organizational structure of gender within the models and corroborate it with linear erasure (\S{\ref{sec:analysis}}).
\end{enumerate}

\section{Method}
\subsection{Linear Erasure}
Given a set of vectors $X=[x_1,\dots,x_n],~x_i\in\mathbb{R}^d$ that associate to a set of target attributes $Z=[z_1,\dots,z_n],~z_i \in \{1,\dots,k\}$, linear erasure entails learning a transformation $f : \mathbb{R}^d \rightarrow \mathbb{R}^d $ such that we can no longer train \textbf{linear classifiers} to predict $z_i$ from $f(x_i)$. When this happens, we say that $Z$ has been linearly erased from $X$. A vital objective in choosing $f$ is to make sure that the transformed embeddings retain all information except the target attribute. 

While there has been several attempts at finding an efficient erasure function~\cite{ravfogel2022rlace,ravfogel2020null}, they do not guarantee erasure or hurt the original embedding space non-trivially. The authors of \cite{belrose2024leace} propose Least Squares Concept Erasure (LEACE), a closed form solution to linear erasure. They begin by proving that if the class centroids of every class in $Z$ coalesce, it implies linear erasure of $Z$ from $X$. They then derive the transformation that ensures class centroid equality with minimal impact to the original embeddings (impact is measured in terms of mean squared norm). They arrive at the LEACE transformation for an embedding $\textbf{x}$ as:
\begin{align}
{f}_{\mathrm{LEACE}}(\textbf{x}) &= \textbf{x} - \mathbf{W}^+ \textbf{P}_{\mathbf{W\Sigma}_{XZ}}\mathbf{W}\big (\textbf{x} - \mathbb{E}[X] \big)
\end{align}
where $\Sigma_{XZ}$ is the covariance between matrices $X$ and $Z$, $\mathbf{W}$ is the whitening transformation $(\Sigma_{XX}^{1/2})^+$ and $W^+$ denotes Moore-Penrose pseudoinverse of matrix $W$. $\textbf{P}_{\mathbf{W\Sigma}_{XZ}} = (\mathbf{W\Sigma}_{XZ})(\mathbf{W\Sigma}_{XZ})^+$ is the orthogonal projection matrix onto $\mathrm{colsp}(\mathbf{W\Sigma}_{XZ})$.


\subsection{Concept Scrubbing}
In \cite{belrose2024leace}, the authors introduce concept scrubbing,  a way to (linearly) remove a selected attribute from \textit{all} layers of a transformer. Concept scrubbing is done by performing linear erasure at the output of each layer in the transformer, before the intermediate representation is fed into the next layer. Scrubbing cannot be done in parallel, since  any intervention at layer $L$ of the transformer affects the hidden states at every subsequent layer $L'>L$. Therefore, linear erasure is done in cascade, beginning at the first layer and sequentially proceeding to the final layer of the model. LEACE makes it easy to implement concept scrubbing since the parameters can be tuned in a streaming fashion, which eliminates the need of storing hidden states in memory or on disk. \vspace{-0.5cm}
\label{sec:conept_scrubbing}
\begin{figure*}[!ht]
    \centering

    \subfloat[Gender Scrubbing for TIMIT]{
        \includegraphics[width=0.48\textwidth,height=5.5cm]{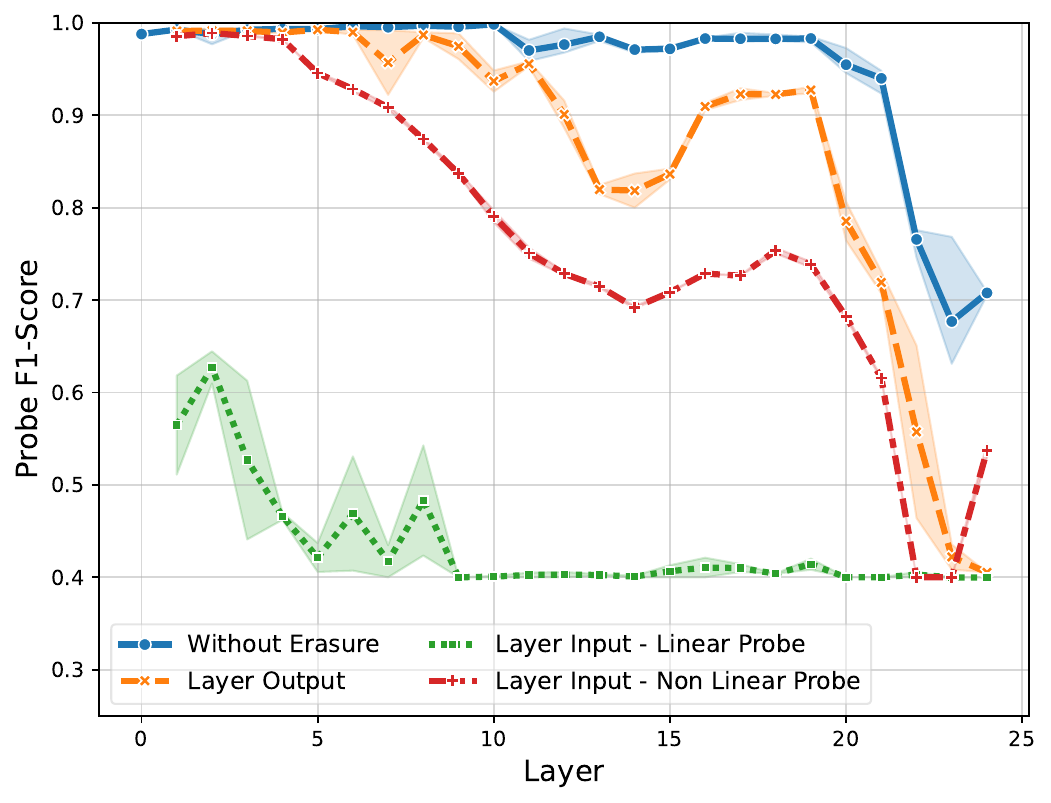}
        \label{fig:scrub_w2v2_timit}
    }
    \hfill
    \subfloat[ Gender Scrubbing for Librispeech-100 ]{
        \includegraphics[width=0.48\textwidth,height=5.5cm]{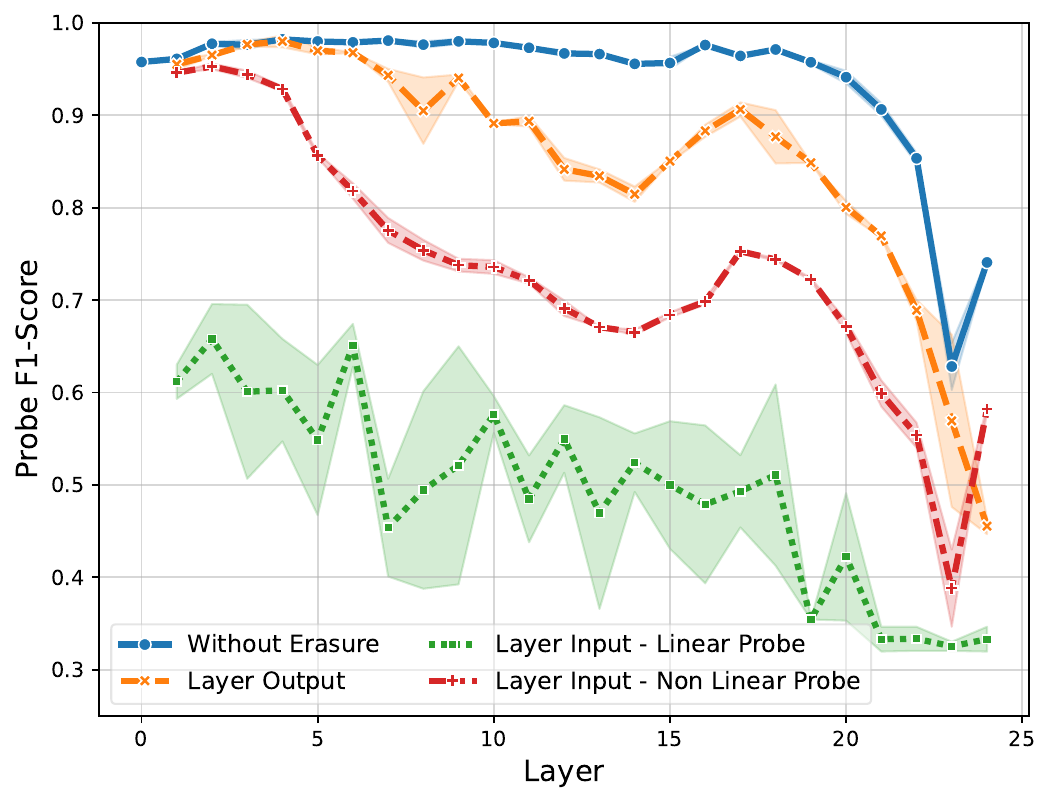}
        \label{fig:scrub_w2v2_librispeech}
    }
    \caption{Gender scrubbing for the Wav2Vec2 ASR model. Plots depict linear probe performances at the input and output when linearly erasing gender from the input at each layer. Mean probe performance on the original model is also shown.}
    \label{fig:scrub_plots}
    \vspace{-0.5cm}
\end{figure*}
\section{Data}
\textbf{TIMIT}~\cite{garofolo1993darpatimit} is a read speech corpus of 630 speakers of various American dialects. Each speaker in the corpus reads 10 phonetically rich sentences in English, and transcripts are made available for ASR applications. TIMIT provides the reported gender of each speaker in the corpus, which is one of Female or Male. 
\textbf{Librispeech}~\cite{panayotov2015librispeech} is a speech corpus of about 1000 hours of read audio-books. Gender is reported for every utterance and is again binary. We use the clean 100 hour training set and the default test set for all experiments. The ASR models used in this paper are both fine tuned using a 960 hour super-set\footnote{There is no test set leakage} of Librispeech. As such, the combination of Librispeech and TIMIT offers an in-domain and out-of-domain scenario for experiments. The speaker and gender distribution for the datasets is shown in Table \ref{table:data}.
\begin{table}[t]
\centering
\begin{tabular}{lcccccc}
    \toprule
    \multirow{2.5}{*}{Dataset} & \multicolumn{2}{c}{\# Utterances}&   \multicolumn{2}{c} {\#Male/\#Female Speakers} \\  
    \cmidrule(lr){2-3} \cmidrule(lr){4-5} 
    & Train & Test &  Train & Test  \\
    \midrule
    TIMIT & 4620  & 1680 & 326/136  & 112/56   \\
    Librispeech & 28539  & 2620 & 126/125  &  20/20   \\
    \bottomrule \\
\end{tabular}
\caption{Dataset statistics for TIMIT and Librispeech. Note that there is no speaker overlap between the train and test sets}
\vspace{-0.75cm}
\label{table:data}
\end{table}
\section{Experimental Setup}
All experiments are run on wav2vec2-large-960h\footnote{\url{https://huggingface.co/facebook/wav2vec2-large-960h}} and hubert-large-ls960-ft\footnote{\url{https://huggingface.co/facebook/hubert-large-ls960-ft}}, which are ASR fine-tuned versions of Wav2Vec2 and HuBERT respectively. Both models have been fine-tuned using the 960 hour Librispeech dataset with a CTC  objective ~\cite{graves2006connectionist}. \textbf{Throughout this work, the models are kept frozen and set to evaluation mode}. The linear probes trained are SGDClassifier models from the $sklearn$~\cite{scikit-learn} library. The non-linear probe trained in Section \ref{sec:gender_scrub} is an multi-layer perceptron classifier from $sklearn$ with default parameters. All probes are trained on three random seeds until the training loss converges. Concept scrubbing is deterministic since the model is frozen and the solution is closed. 

\section{Gender Scrubbing}
\label{sec:gender_scrub}
In this section, we employ the concept scrubbing method described in Section \ref{sec:conept_scrubbing} across the layers of both ASR models. Let us start by assuming an audio corpus $\textbf{X}$. Each audio sample $x_i \in X$ is associated with a gender label $z_i \in \{Female,Male\}$. When $x_i$ is fed into an ASR transformer, each encoder layer $L_j$ produces $H$ dimensional latent representations across $T_i$ time steps of $x_i$, generating an embedding of size $(T_i,H)$. We take the mean across the temporal dimension $T$ to obtain a single  representation for each utterance.  Then, we train linear classifier to predict the target attribute from the mean-pooled utterance representations. We call this \textbf{Mean Probing}. The performance of the probe is used as a measure of the gender information available at this layer. To erase gender information at layer $L_j$, we train a LEACE eraser $E_j$ over $T_i$ embeddings using $z_i$ as target label. The eraser is trained over every data point in $X$ in a similar fashion.  Once training finishes, we loop through the entire dataset again, erasing the gender information using the trained $E_j$. These \textit{gender-erased} embeddings are now fed into the next layer $L_{j+1}$, and the process is repeated until the last transformer layer. By erasing gender sequentially from one end to the other, we ensure that no layer of the transformer network has access to linearly encoded gender.

\subsection{Tracking Erasure}
\label{sec:tracking_erasure}
To monitor the traversal of encoded gender along the model, linear probes are trained at the input and the output of each layer. A performance difference between the input/output probes at any layer can then be attributed to the operations of the layer itself. Note that the input at each layer is the gender-erased output from the previous layer. The input/output probe performances for TIMIT and Librispeech-100h (using Wav2Vec2) is visualized\footnote{\label{github_footnote}The reader is encouraged to view complete set of plots in this link: \url{https://github.com/Krishnan-Aravind/Interspeech_2024_Gender}} in Figure \ref{fig:scrub_plots}. For both datasets, we see that the linear probes trained at the input have a near-random performance, which suggests that the erasure operation is successful at each layer. However, the results for the output probe show that the model recovers gender almost entirely in the early layers. This recovery can be attributed to two factors: (1) the non-linearity available to the transformer layer, and (2) the parametric capacity of the layer. To isolate the contribution of each of these factors, we train a simple non-linear probe on the linearly-erased input embeddings. The absolute performance of the non-linear classifier reveals any non-linearly encoded gender that lingers after linear erasure. The performance gap between the non-linear input probe and the output probe is attributed to the parametric capacity of the layer. 

The performances of the non-linear probe show that the erased representations still contain non-linearly encoded gender, especially in the early layers of the model. However, non-linearly encoded gender in the scrubbed inputs and the consequent recovery of gender information is seen to weaken in the latter layers. 
 The last two layers in particular are unable to recover the erased gender information at all. All linear probes trained on these layers show near-random performance, suggesting a complete linear erasure of gender information in these layers. Note that all operations of the model after the last transformer layer are linear so any non-linearly encoded gender present in the last layer's output is discarded in the subsequent stage. We conclude that gender can be linearly erased without recovery from the latter layers of Wav2Vec2 and HuBERT\footnotemark[4]. 

\subsection{Effects on downstream ASR}
\label{sec:downstream_asr}
 \vspace{-0.1cm}For performing ASR with Wav2Vec2 or HuBERT, representations from the last encoder layer is fed into a linear layer (called the \textit{language modelling head}) which then classifies it into a speech symbol.  Following gender scrubbing, we feed the gender-erased embeddings from the last layer into the language head and measure WER at the output. Note that the language head is a linear layer, so linear erasure at the last layer masks gender from this layer as well.  The WER obtained after scrubbing is compared with the original WER obtained when we do not intervene in the transformer's layers. This comparison sheds light on two questions: (1) what are the consequences of erasing gender on the intermediate subspace of the ASR model?  and (2) is gender information in the last layers utilized during ASR despite its "presence" in these layers? The comparison is tabulated in Table \ref{table:scrubbing_wer}. In most cases, we see that gender scrubbing has a slight effect, albeit a noticeable drop on the WER of the model. The results suggest a reduced overlap between the linguistic and (linear) gender subspaces in the last layer, since we would have seen a more prominent drop in WER if the eraser was interfering with the linguistic space when removing gender.\vspace{-0.1cm}
\begin{figure*}[!ht]
    \centering    
    \subfloat[Snapshot probe performance: Wa2Vec2]{
        \includegraphics[width=0.32\textwidth]{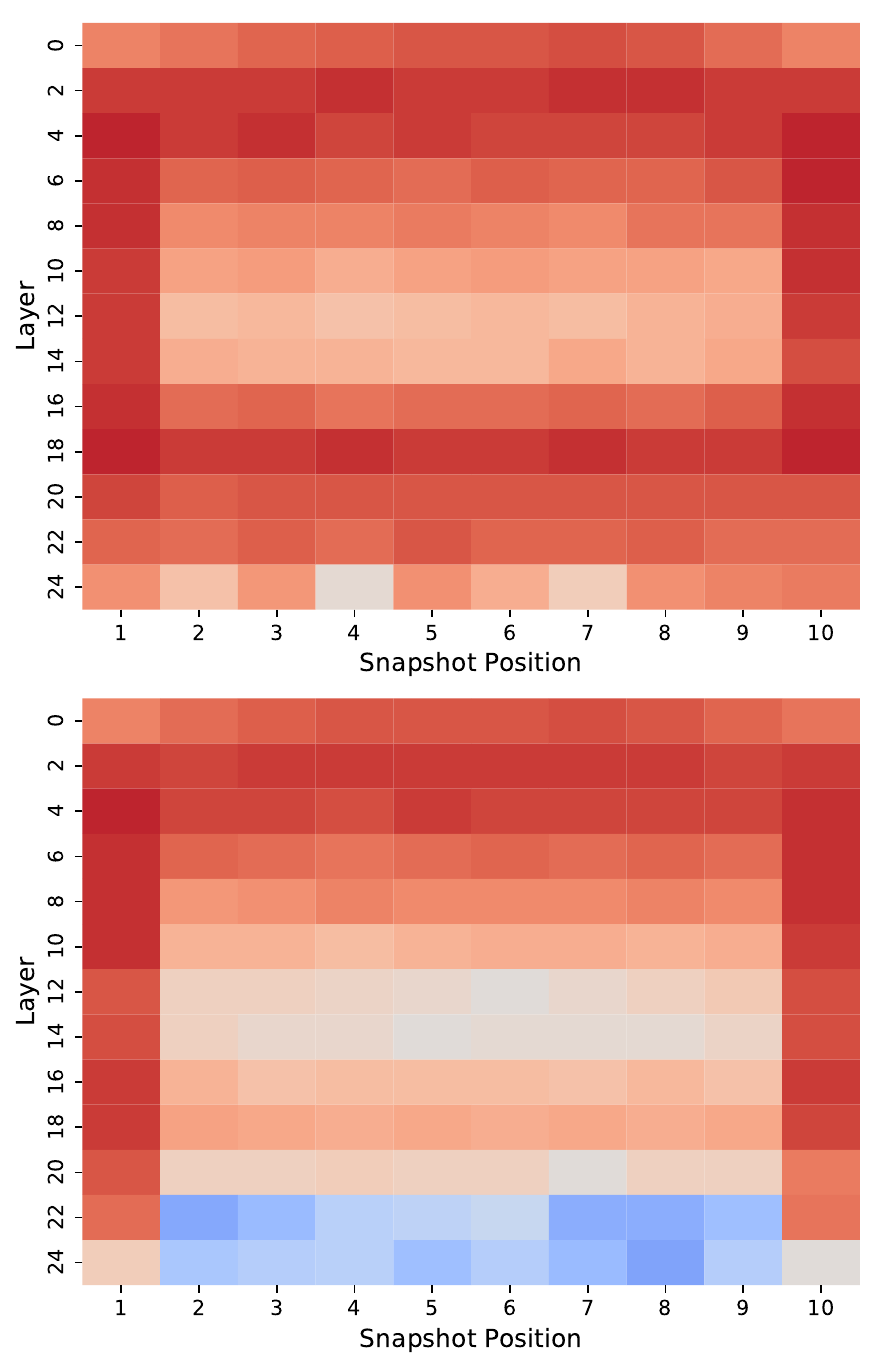}
        \label{fig:snapshot_layers_Wav2Vec2}
    }
    \subfloat[Snapshot probe performance: HuBERT]{
        \includegraphics[width=0.32\textwidth]{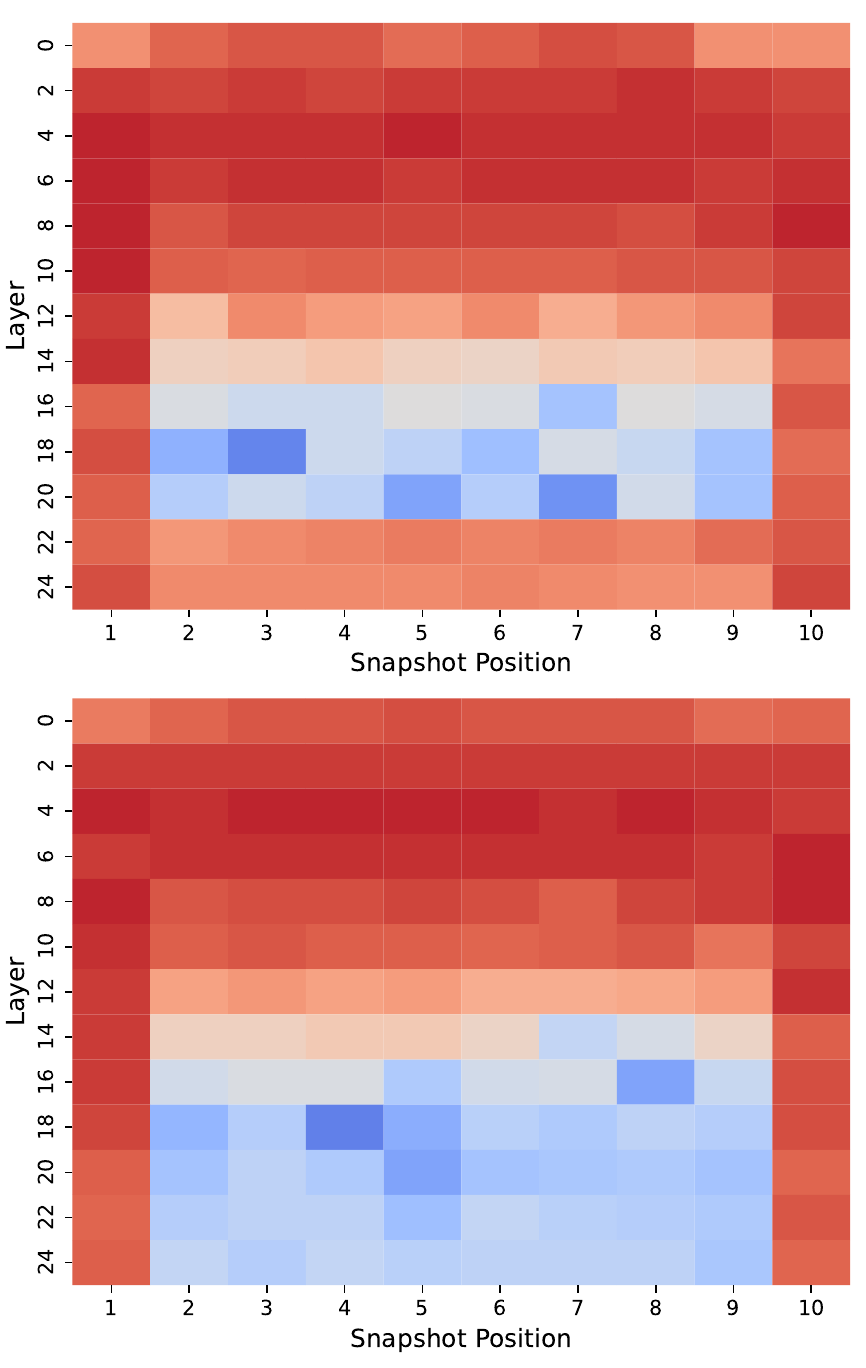}
        \label{fig:snapshot_layers_HuBERT}
    }
    \subfloat[ Cross-positional snapshot probing at the final layer of the Wav2Vec2 models]{
        \includegraphics[width=0.32\textwidth,height=8.6cm]{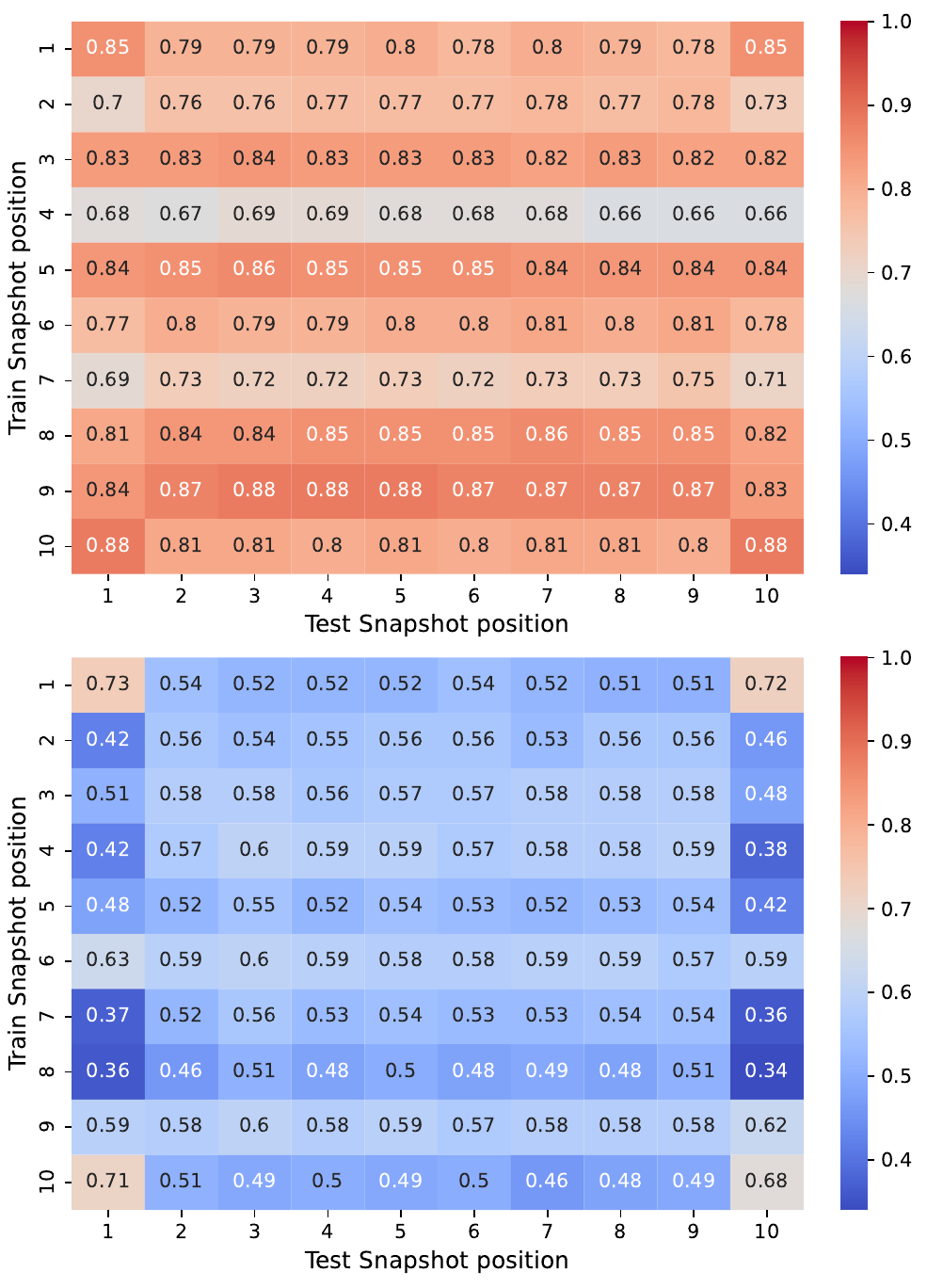}
        \label{fig:snapshot_l24}
    }
    \caption{Snapshot probing on the layers of pretrained (above) and fine-tuned (below) Wav2Vec2 and HuBERT. The colors indicate the F1-score of a probed trained at the indicated position. Results are shown for Librispeech.}
    \vspace{-0.3cm}
    \label{fig:snapshot_probes}
\end{figure*} 
\begin{table}[b]
\centering
\begin{tabular}{lcccccc}
    \toprule
    \multirow{2.5}{*}{Dataset} & \multicolumn{2}{c}{Wav2Vec2}&   \multicolumn{2}{c}{HuBERT}  \\
    \cmidrule(lr){2-3} \cmidrule(lr){4-5} 
    & Original & Scrubbed &  Original & Scrubbed  \\
    \midrule
    TIMIT & 23.96  & 24.18 & 19.23  & 21.57   \\
    Librispeech & 2.76  & 2.77 & 2.07  &   2.90  \\
    \bottomrule \\
\end{tabular}
\caption{WER before and after gender scrubbing}
\label{table:scrubbing_wer}
\vspace{-1cm}
\end{table}
\section{ Analysis }
\label{sec:analysis}
In Figure \ref{fig:scrub_plots}, we observe that the probe trained without intervention shows comparable performances between layer 10 and layer 21 but the probe trained after linear erasure behaves very differently in these layers. In the analysis section, we ask the following question: ``Why is non-recoverable gender removal more successful in the final layers despite comparable original probe performances to the initial layers?''. To address this question,  we investigate how gender is organized at a frame-level within the hidden representations.
\subsection{Frame-level probing}
\vspace{-0.1cm}
In Section  \ref{sec:gender_scrub}, we trained linear probes on mean pooled layer representations. Henceforth, we will train probes on a frame-level, meaning we train linear probes on the individual $(T,H)$ representations at each layer. This presents one hurdle: each audio signal $\textbf{x}_i$ has a different number of time steps and produces an embedding of shape $(T_i,H)$ at each layer. For efficient comparisons, we take 10 equally spaced snapshots across time steps, starting with the first and ending with the last. This forces each layer-wise representation to $(10,H)$ dimensions. We can now train probes on any one of these snapshots, using the same gender label $z_i$ across all snapshots of $\textbf{x}_i$: we call this \textbf{Snapshot probing}. Initially, we train 10 snapshot probes at each layer, training and testing on the \textit{same} snapshot position. The performances of these snapshot probes for the layers of the pretrained and fine-tuned Wav2Vec2 and HuBERT (using Librispeech\footnotemark[4]) is shown in Figures \ref{fig:snapshot_layers_Wav2Vec2} and \ref{fig:snapshot_layers_HuBERT}. Across the layers of the pretrained models, we see that the probes perform strongly over almost all snapshot positions. Trends in the middle layers resemble the behaviour of an autoencoder, where gender information localizes in the ends. This is consistent with the findings reported in \cite{pasad2021layer}, where the authors observe an autoencoder-like tendency within self supervised speech models.  Fine-tuning is seen to strongly affect the last layers of the model, where we see an overall reduction in gender-awareness, accompanied by a strong localization of gender into the first and last snapshots (which correspond to the first and last frames of the audio files respectively). In the last four layers of Wav2Vec2 and HuBERT, the performances of the intermediate snapshots fall near-random, suggesting a complete absence of linearly encoded gender in these positions. Fine-tuning could be removing gender from the central snapshots in the final layers to make way for linguistic information that facilitates the ASR task. Note that the pretrained HuBERT shows this localization strategy during pretraining already; The final layers additionally \textit{clear up} after fine-tuning. 

In Figures \ref{fig:snapshot_layers_Wav2Vec2} and \ref{fig:snapshot_layers_HuBERT}, we also see that the snapshot probes trained at the first and last positions match/outperform the other positions after layer 0 (which is the input to the first transformer layer). This is intriguing, since these positions often contain silence and therefore cannot naturally encode speaker characteristics. The simplest explanation is that the model might be re-purposing these positions to store auxiliary information, utilizing the cross attention mechanism to redirect information into them.  Frame level probing reveals a heterogeneous frame-level organization of gender across the layers of an ASR transformer, despite the exhibition of consistent mean probe performances.\vspace{-0.1cm}
\subsection{Cross-Position Snapshot Training}
\vspace{-0.1cm}
Figures \ref{fig:snapshot_layers_Wav2Vec2} and \ref{fig:snapshot_layers_HuBERT} depict snapshot probes trained and tested at the \textit{same} positions in each layer. Alternatively, snapshot probes can be trained and tested on \textit{different} positions within the same $(T_i,H)$ embedding. A high probe performance between two positions then indicates a shared gender-subspace between them. For the Wav2Vec2 model, results from training and testing snapshot probes at various positions in the final layer is shown in Figure \ref{fig:snapshot_l24}. Note that the diagonals of the cross-position plots in Figure \ref{fig:snapshot_l24} correspond to the last rows in Figure \ref{fig:snapshot_layers_Wav2Vec2}. For the pretrained model, we see high transferability across most snapshot positions, which suggests the existence of a shared gender-subspace across frame-level representations in this layer. Trends are different in for the fine-tuned model, where we see a widespread purge of gender from all snapshots except the first and the last. We also see that the probes trained on these positions have a high transferability, which suggests that the gender-subspace is shared only between these positions in the fine tuned model. Clearly, fine tuning re-purposes the gender-subspace in the last layer to store linguistic information in all frames except the first and the last. This adds to our previous conclusions : In the final layer, gender information is stored in the first and last frame positions in a \textit{ shared subspace}. 

The findings in Figure \ref{fig:snapshot_probes} help interpret the trends seen in Figure \ref{fig:scrub_plots}: We can remove gender with reduced recovery from the last layers because it is localized and stored in shared subspaces in these layers, while removal from the initial layers is difficult owing to the omnipresence of gender in these layers. \vspace{-0.1cm}
\section{Conclusions and an Ethical Note}
\vspace{-0.1cm}
This work explores the utilization of linearly detectable gender in two transformer-based ASR models, Wav2Vec2 and HuBERT. We first employ linear erasure to remove gender from each layer of the model and find that it can be removed from the final layers without  damaging the ASR performance heavily. Our experiments demonstrate the feasibility of creating gender-neutral ASR embeddings in the linear space. To explain the ease of gender removal from the latter layers, we explore the frame-level organization of gender in the model's hidden representations. We find that in the latter layers, gender-information concentrates in a shared subspace within the embeddings of the initial and final positions. This facilitates the removal of gender cues from the final layers of the model. Our work sheds light on how ASR models encode gender and pave way to creating usable but speaker-blind  embeddings.

We endorse the ethical concerns posited by \cite{attanasio2024multilingual} and do not make gender inferences apart from those declared, or endorse normativity. Our use of \textit{gender} in-place of  \textit{gender binary} emerges from the labels made available in the datasets we use. We simply study \textit{how} the models in question encode and use the gender categories reported in the datasets and any mention of gender should be interpreted accordingly. We consider the lack of non-binary labels a limitation of this work.

\section{Acknowledgements}
We thank the anonymous reviewers for their constructive feedback. We also thank Marius Mosbach, Dawei Zhu, Jesujoba Alabi, Vagrant Gautam and Bernd Möbius for their suggestions. This research is funded by the Deutsche Forschungsgemeinschaft, Project-ID 232722074 -- SFB 1102, and the ViCKI project of the Federal Ministry for Economic Affairs and Climate Action (BMWK) under grant number FKZ 20D1910D.


\bibliographystyle{IEEEtran}
\bibliography{mybib}

\end{document}